%% file: arxive.tex
\documentclass[runningheads]{llncs}
\usepackage[T1]{fontenc}
\def\our{MedGS}
\usepackage{graphicx,verbatim}
 \usepackage{booktabs}
\usepackage{graphicx}
\usepackage{amsmath}
\usepackage{amssymb}
\usepackage{multirow}
\usepackage{subcaption}
\usepackage{wrapfig}
\usepackage{booktabs} 

\def\m{\mathrm{m}}

\def\x{\mathrm{x}}

\def\s{\mathrm{s}}

\def\our{MedGS}
\def\ourinter{MedGS$_{i}$}
\def\ourseg{MedGS$_{s}$}

\def\reg{IBFR}

\usepackage[utf8]{inputenc}
\usepackage[T1]{fontenc}
\usepackage[table]{xcolor}
\definecolor{best}{RGB}{255,220,220}

\begin{document}
\title{\our{}: Learning 3D Structures from Sequential Medical Imaging with Gaussian Splatting}
\titlerunning{\our{}}
%
\author{Kacper Marzol\inst{1} \and
Ignacy Kolton\inst{1} \and
Weronika Smolak-Dy\.zewska\inst{1} \and
Joanna Kaleta\inst{2,4} \and
\.Zaneta \'Swiderska-Chadaj\inst{2,3} \and
Marcin Mazur\inst{1} \and
Miros{\l}aw Dziekiewicz\inst{5} \and
Tomasz Markiewicz\inst{2,5} \and
Przemys{\l}aw Spurek\inst{1,3}  
}
\authorrunning{K. Marzol et al.}
%
\institute{ Jagiellonian University, Poland \and
Warsaw University of Technology, Poland
\and
IDEAS Research Institute, Poland \and
Sano Centre for Computional Medicine, Poland \and
Military Institute of Medicine, Poland
}

  
\maketitle              
\vspace{-0.7cm}

\begin{figure}
\includegraphics[width=0.95\textwidth]{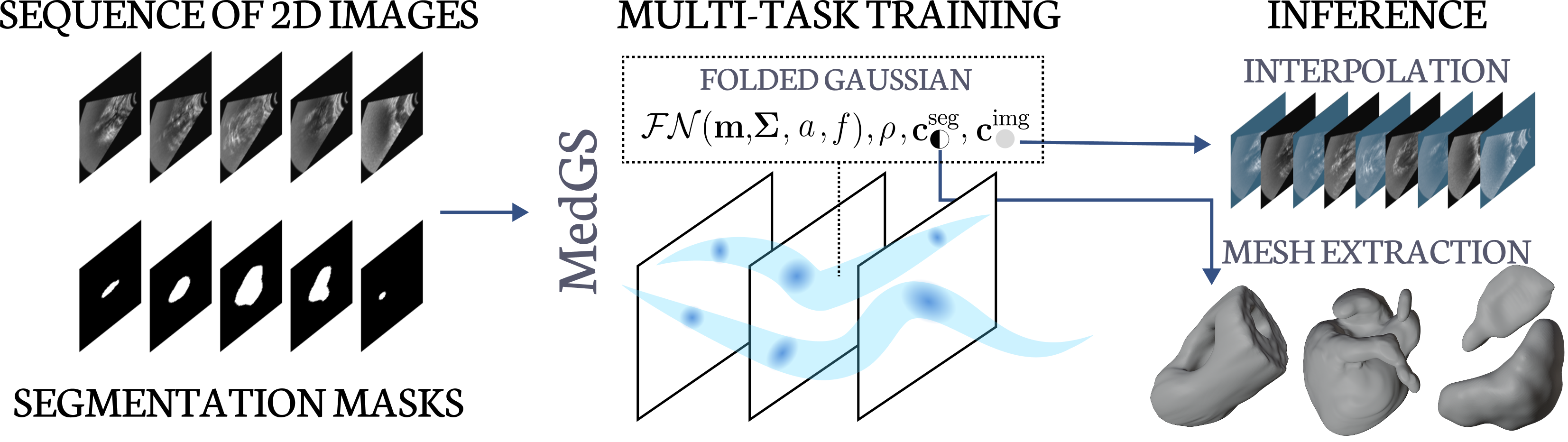}
\caption{Our \our{} model shares a unified geometry to simultaneously solve frame interpolation and mask reconstruction tasks. This multi-task coupling leverages the visual signal from the image branch to regularize the underlying Folded-Gaussian distribution. This synergy ensures geometric continuity and enables high-fidelity 3D mesh extraction even from sparse input slices.
} 
\label{fig:teaser}
\end{figure}
\vspace{-1.1cm}

\begin{abstract}

Multi-modal three-dimensional (3D) medical imaging data, derived from ultrasound, magnetic resonance imaging (MRI), and computed tomography (CT), provide a widely adopted approach for non-invasive anatomical visualization. However, accurate modeling depends on surface reconstruction and frame-to-frame interpolation,\linebreak where traditional methods often struggle with image noise and incomplete information between sparse frames. To address these challenges, we present \our{}, a novel framework based on Gaussian Splatting (GS) designed for high-fidelity 3D anatomical reconstruction. 
Uniquely, \our{} employs a multi-task architecture that simultaneously performs frame interpolation and segmentation using a unified geometric representation. By coupling these tasks, the model leverages dense signals from image synthesis to regularize the geometry, enabling high-quality surface extraction even from a limited number of input frames. 
Specifically, medical data are modeled as Folded-Gaussians with dual color attributes, supported by an In-Between Frame Regularization (IBFR) mechanism. Experimental results demonstrate that \our{} achieves higher metric scores than implicit neural representations and improves interpolation quality.
The source code is available at 
\url{https://github.com/gmum/MedGS}.


\end{abstract}
%
%
%

\section{Introduction}
\label{sec:intro}

Multi-modal three-dimensional (3D) medical imaging, including ultrasound, MRI, and CT \cite{guan2009study}, is essential for non-invasive anatomical visualization. A~central objective in this field is high-fidelity surface reconstruction for surgical planning and analysis. While classical methods like marching cubes extract meshes from voxel grids, they often produce connectivity artifacts and remain limited by resolution and segmentation fidelity \cite{mohamed2019survey}.

3D reconstruction in this context typically uses sequential 2D planar cross-sections embedded in 3D space. This setting differs fundamentally from tomographic reconstruction \cite{cai2024radiative,cai2024structure,lin2025pixel,zha2024r2}, which resolves line integrals from projection views. Planar reconstruction, in contrast, focuses on frame-to-frame interpolation and geometric continuity between slices.

Recent advances in deep learning, particularly Implicit Neural Representations (INRs), have demonstrated high-fidelity reconstruction for planar modalities \cite{grutman2025implicit,sitzmann2020implicit,wysocki2024ultra,xu2023nesvor}. Despite their accuracy, INRs are computationally intensive and difficult to interpret or manually edit, which poses a significant limitation for deformable registration and computer-assisted surgery \cite{figl2019deformable,wang2025narrative}. Hybrid approaches \cite{liu2025medical} partially improve efficiency but retain neural complexity. Conversely, pure Gaussian Splatting (GS) \cite{kerbl20233d} offers an explicit, efficient representation enabling direct geometric manipulation. This is critical for recovering morphologically consistent structures from deformed scans \cite{hoffmann2021synthmorph,jiang2023defcor}, a capability largely absent in implicit models.

\begin{figure}[t]
  \centering
  \includegraphics[width=0.90\textwidth]{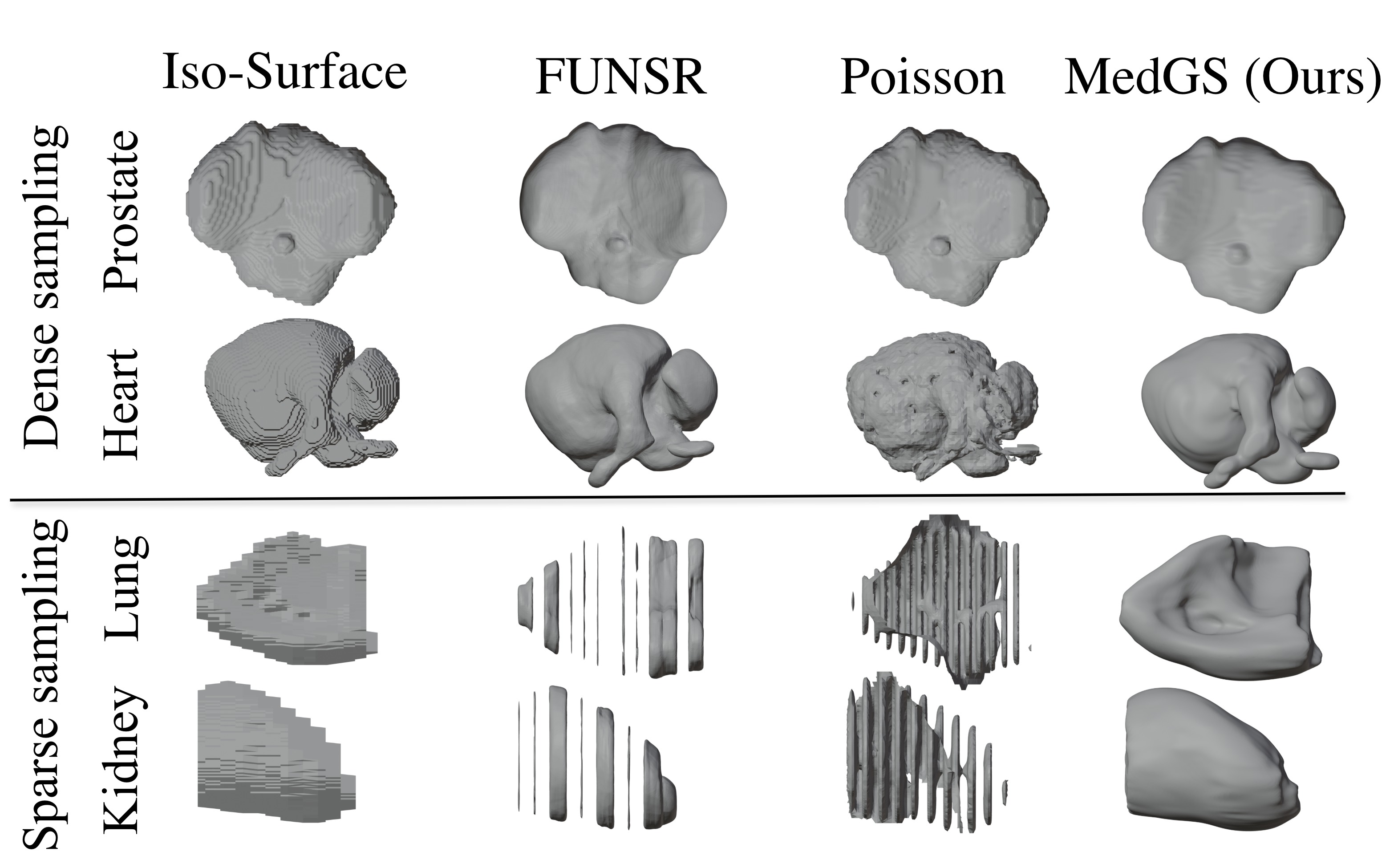}
  \caption{Qualitative mesh reconstruction results on prostate ultrasound data from \cite{zachary_m_c_baum_2023_7870105}, and on heart, lung, and kidney from \cite{d2024totalsegmentator}.  Our method demonstrates superior topology preservation, edge continuity, and overall surface geometry. FUNSR and Poisson produce less smooth meshes and fail under sparse sampling conditions.}
  \label{fig:prostate_mesh}
\end{figure}

In this paper, we introduce \our{}, a GS-based framework for high-fidelity 3D anatomical mesh reconstruction. \our{} represents volumetric data as sequences of 2D cross-sections in 3D space. It uses a multi-task design where Gaussian primitives share a unified geometry, including positions and temporal deformations,  
while separate color branches handle
frame interpolation and segmentation. This shared-geometry, dual-appearance setup
enables simultaneous grayscale regression and mask-based surface extraction, leveraging dense signals from image synthesis to regularize geometry.
As a result, \our{} delivers high-quality segmentation and smooth mesh reconstruction even from sparse frame sequences, where decoupled methods often fail due to a lack of geometric continuity.
Building upon VeGaS \cite{smolak2024vegas} and Folded-Gaussian distributions, it captures complex nonlinear structures while retaining the efficiency and interpretability of explicit representations.
Our contributions are summarized as follows:
\begin{itemize}
    \item We propose \our{}, a Gaussian Splatting-based framework tailored for high-fidelity 3D organ mesh reconstruction from medical imaging data, enabling accurate modeling of complex anatomical structures.
    \item We introduce a multi-task learning strategy that regularizes and improves segmentation-driven surface reconstruction in sparse or noisy settings.
    \item We demonstrate that MedGS provides accurate results, enabling clear visualization and tracking of anatomical changes over time. This supports reliable longitudinal monitoring of pathologies such as aneurysm expansion and renal tumor growth, offering valuable insight into disease progression. 
\end{itemize}

\section{Related Works}

Traditional 3D medical reconstruction methods often rely on direct extraction \cite{kerr2017accurate,zhang2004surface,zhang2002direct}, but they are sensitive to noise and incomplete data. Deep learning has improved robustness using Graph Convolutional Networks for mesh deformation \cite{nakao2021image} and end-to-end frameworks for joint segmentation and reconstruction \cite{gopinath2021segrecon,zhou2019handbook}. In domains like brain MRI and echocardiography, deformable templates and weakly supervised autoencoders further enhance shape modeling \cite{bongratz2024neural,laumer2023weakly,ma2022cortexode}.  

Implicit Neural Representations (INRs) achieve high-fidelity reconstruction in MRI \cite{xu2023nesvor}, CT \cite{song2023piner}, and ultrasound \cite{grutman2025implicit,wysocki2024ultra}, often via Signed Distance Functions \cite{ma2020neural,park2019deepsdf} or occupancy functions \cite{amiranashvili2024learning,chen2024neural}, but remain computationally expensive and hard to edit.  

Gaussian Splatting (GS) has recently been adapted for medical imaging, e.g., radiative GS for sparse X-ray/CT \cite{cai2024radiative,lin2025pixel,zha2024r2} and hybrid tri-plane GS for segmentation \cite{liu2025medical}. In contrast, \our{}, built on VeGaS~\cite{smolak2024vegas}, targets sequential planar data with explicit GS, shared geometry, and dual-color architecture, enabling efficient modeling, robust mesh extraction, and direct editability of Gaussian primitives.

\section{Preliminaries}  

\our{} builds upon two foundational models: Gaussian Splatting (GS) \cite{kerbl20233d} and Video Gaussian Splatting (VeGaS) \cite{smolak2024vegas}. We summarize both below.  

Gaussian Splatting (GS)~\cite{kerbl20233d} is an explicit 3D scene representation that models a scene as a collection of Gaussian primitives $(\mathcal{N}(\mathbf{m}, \mathbf{\Sigma}), \rho, c)$. Each Gaussian is characterized by its mean (position) $\mathbf{m}$, an anisotropic covariance (shape and orientation) $\mathbf{\Sigma}$, an opacity $\rho$, and view-dependent color $c$ represented using spherical harmonics.


\begin{figure}[t!]
  \centering
  \includegraphics[width=0.99\textwidth]{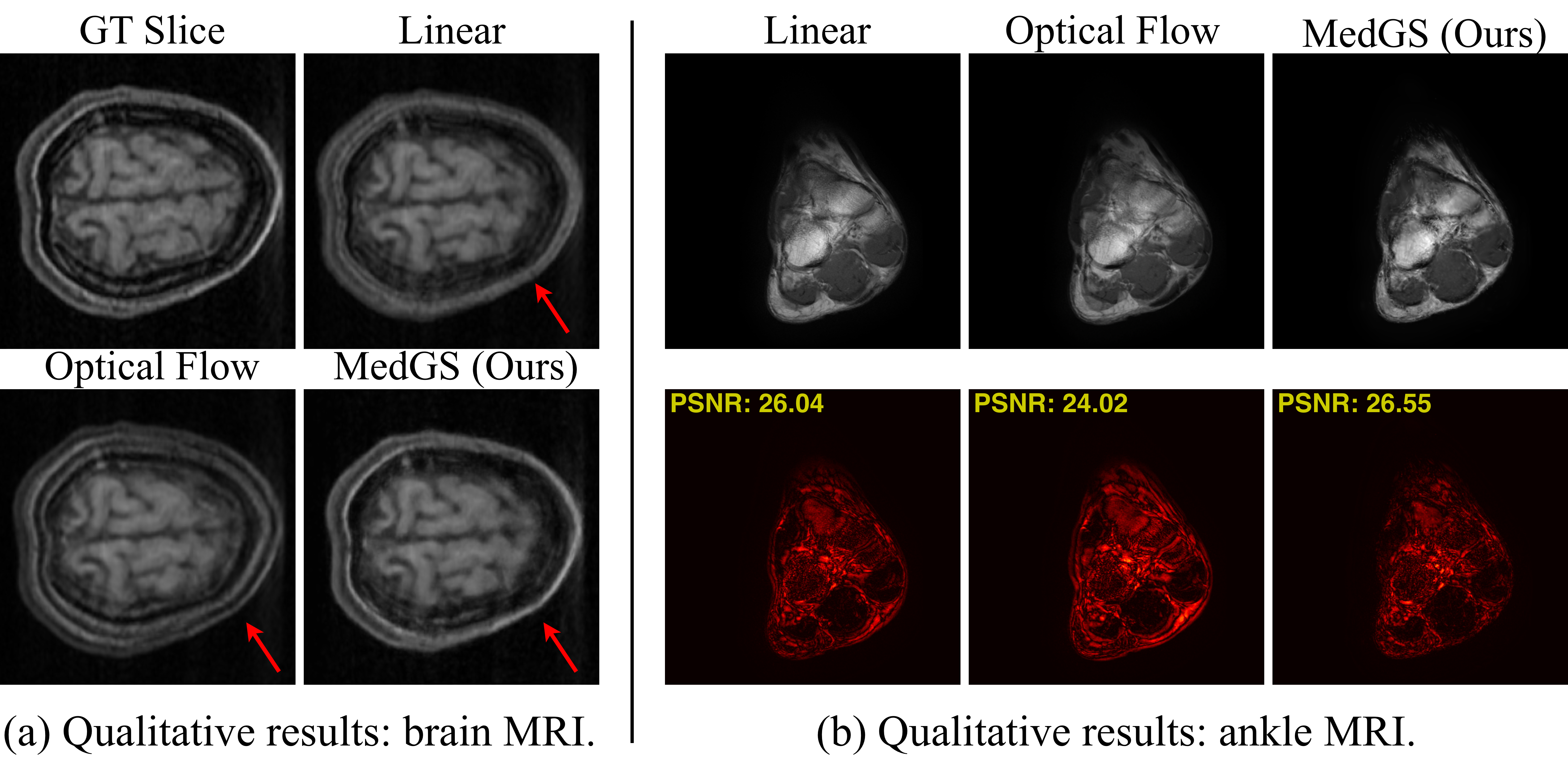}
  \caption{Qualitative results of frame interpolation on MRI data. \our{} produces sharper reconstructions. For the ankle MRI, the MSE is also visualized using a heat map.}
  \label{fig:interp}
\end{figure}

Video Gaussian Splatting (VeGaS)~\cite{smolak2024vegas} extends GS to dynamic sequences by introducing Folded-Gaussian distributions, which better capture temporal and nonlinear spatial changes. In VeGaS, video frames are represented as parallel planes in 3D, and each scene element is modelled as a 3D Folded-Gaussian conditioned on frame timestamps. Temporal dynamics are encoded via learnable functions applied to the mean $\mathbf{m}=(\m_s, \m_t)$ and covariance matrix $\mathbf{\Sigma}=\text{diag}(\mathbf{\Sigma}_\s,\sigma^2_t)$, allowing flexible shifts and rescaling over time:  
\begin{equation}\label{eq:cond1}
\mathcal{FN} (\mathbf{m}, \mathbf{\Sigma}, a, f)(\mathbf{x}) = \mathcal{N}(\m_s+f(\m_t-t),a(t)\mathbf{\Sigma}_\s) (\s|t) \cdot \mathcal{N}(\m_t, \sigma_t^2)(t),
\end{equation}  
where \(\s|t\) denotes the spatial variable conditioned on time \(t\), \(\x=(\s|t,t)\), and \(a(t)\), \(f(t)\) encode temporal rescaling and nonlinear shifts. This design efficiently captures both persistent features (via broad Gaussians) and transient events (via localized Gaussians). The explicit representation enables accurate frame interpolation, dynamic scene reconstruction, and interactive editing. VeGaS thus overcomes limitations of static GS and implicit neural representations for dynamic data, making it well-suited for spatiotemporal medical sequences, where frames occur at fixed intervals and noise or missing data are common~\cite{smolak2024vegas}.

\section{\our{}}
\label{sec:method}

\begin{figure}[t]
  \centering
  \includegraphics[width=0.99\textwidth]{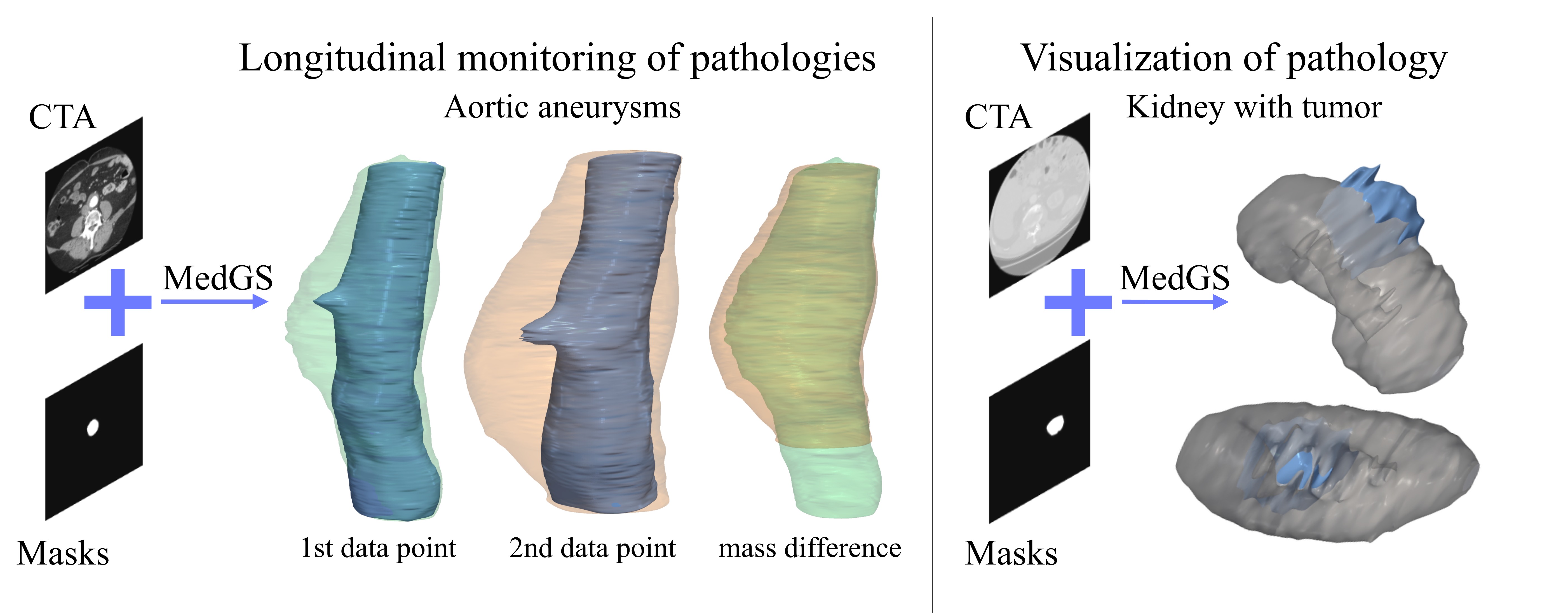}
  \caption{CTA-based 3D reconstructions for advanced disease monitoring. Left: Longitudinal tracking of an abdominal aortic aneurysm. Mass in 1st and 2nd data point is colored in light green and light orange, respectively. Right: Precise localization of a renal tumor within the kidney's cortex and medulla.}
  \label{fig:kidney_cancer}
\end{figure}

\our{} is a unified Gaussian Splatting-based framework designed for the simultaneous reconstruction of anatomical appearance and surface geometry. Unlike traditional approaches that treat image interpolation and segmentation as disparate tasks, \our{} integrates them into a single coherent model. By sharing the underlying geometric parameters while maintaining distinct appearance modules, the model ensures that the reconstructed organ surfaces are anatomically consistent with the interpolated medical imagery.

\noindent \textbf{Unified Geometric Representation}
The core of \our{} is a collection of 3D Folded-Gaussians that share spatial and temporal attributes but possess dual appearance characteristics. Formally, the model is defined as a set of primitives:
\begin{equation}
\label{eq:medgs_unified}
\mathcal{G}_{\text{\our{}}} = \left\{ \left( \mathcal{FN}(\mathbf{m}_i, \mathbf{\Sigma}_i, a_i, f_i), \rho_i, \mathbf{c}_i^{\text{img}}, \mathbf{c}_i^{\text{seg}} \right) \right\}_{i=1}^N,
\end{equation}
where $N$ is the number of Gaussians.
Here, the geometry is parameterized by the Folded-Gaussian distribution $\mathcal{FN}$ (defined in Eq.~\ref{eq:cond1}), comprising the mean $\mathbf{m}_i$, covariance $\mathbf{\Sigma}_i$, and temporal deformation functions $a_i, f_i$. The opacity is denoted by $\rho_i$.
Crucially, each Gaussian $i$ carries two distinct sets of color coefficients:

\begin{itemize}
    \item $\mathbf{c}_i^{\text{img}}$: Reconstructs the grayscale intensity of the medical scan.
    \item $\mathbf{c}_i^{\text{seg}}$: Reconstructs the binary occupancy mask of the anatomical structure.
\end{itemize}
This shared parameterization enforces a strong structural prior: the geometric primitives defining the organ's boundary in the segmentation task are the same primitives generating the visual texture in the interpolation task.



\noindent \textbf{Simultaneous Dual-Task Optimization}
Training \our{} involves optimizing the shared geometry and both color sets simultaneously. The input data consists of tuples $(I_t, M_t, t)$, where $I_t$ is the grayscale image, $M_t$ is the binary mask, and $t \in [0,1]$ is the normalized timestamp.

\noindent {\em In-Between Frame Regularization (\reg{})}
The first head of the model, utilizing $\mathbf{c}^{\text{img}}$, focuses on reconstructing the medical frames. Medical imaging data often suffers from noise and sampling sparsity. To mitigate this, we employ \reg{} - \textit{In-Between Frame Regularization}.
For consecutive frames $I_t$ and $I_{t+1}$, we generate a pseudo-ground-truth interpolated frame:
$
I_{t_{\alpha}} = \alpha \cdot I_{t} + (1-\alpha) \cdot I_{t+1},
$
where $\alpha \sim \mathcal{U}(0.2, 0.8)$. These synthetic frames introduce a regularization loss, $\mathcal{L}_{\text{interp}}^{\ast}$, that forces the Folded-Gaussians to smoothly approximate transitions between slices, thereby preventing overfitting to noise or collapsing into discrete layers.

\noindent {\em Mesh Reconstruction via Mask Field}
The second head, utilizing $\mathbf{c}^{\text{seg}}$, learns the volumetric shape of the organ. Since segmentation masks $M_t$ are binary and typically noise-free, \reg{} is not applied to this branch. Instead, the model learns to render a continuous 3D scalar field, in which the rendered value approximates the probability of occupancy. Once trained, we extract the high-fidelity 3D mesh by querying the model at dense temporal intervals to produce a super-resolved mask volume, which is then processed with the Marching Cubes algorithm.










\begin{table}[t]
\centering
\caption{Mesh reconstruction results on Prostate dataset - Chamfer Distance (CD) and 95th Percentile Hausdorff Distance (HD95) for specimens 65-70.}
\label{tab:prostate_mesh_combined}
\fontsize{8pt}{9pt}\selectfont{
\begin{tabular}{@{}l
  cccccc |  
  cccccc  
  @{}}
\toprule
Method
& \multicolumn{6}{c}{CD $\downarrow$}
& \multicolumn{6}{c}{HD95 $\downarrow$} \\
& 65 & 66 & 67 & 68 & 69 & 70 
& 65 & 66 & 67 & 68 & 69 & 70 \\
\midrule
Baseline 
& 0.248 & 0.206 & 0.216 & 0.214 & 0.219 & 0.236 
& 0.425 & 0.353 & 0.374 & 0.369 & 0.378 & 0.411  \\
FUNSR
& 0.239 & 0.216 & 0.227 & 0.217 & 0.209 & 0.249 
& 0.421 & 0.384 & 0.393 & 0.374 & 0.374 & 0.426  \\
Poisson
& 0.235 & 0.204 & 0.186 & 0.198 & 0.198 & 0.225 
& 0.397 & 0.361 & 0.334 & 0.345 & 0.355 & 0.400  \\
\ourseg{}
& \textbf{0.194} & 0.207 & 0.179 & \textbf{0.177} & \textbf{0.187} & 0.212 
& \textbf{0.365} & 0.350 & 0.325 & 0.332 & \textbf{0.350} & 0.386 \\
\our{}
& \textbf{0.194} & \textbf{0.184} & \textbf{0.172} & \textbf{0.177} & 0.186 & \textbf{0.208} 
& \textbf{0.365} & \textbf{0.333} & \textbf{0.318} & \textbf{0.329} & \textbf{0.350} & \textbf{0.384}  \\
\bottomrule
\end{tabular}%
}
\end{table}

\begin{table}[t]
\centering
\begin{minipage}{0.49\textwidth} 
\centering
\caption{Ablation study of \ourinter{} on MRI brain interpolation using every 2nd frame. 
Reported metrics include PSNR and SSIM.}
\label{tab:ablation}
\fontsize{8pt}{9pt}\selectfont
\begin{tabular}{lcc}
\toprule
Variant & PSNR $\uparrow$ & SSIM $\uparrow$ \\
\midrule
w/o $\mathcal{L}_{\text{interp}}^{\ast}$ & 32.68 $\pm$ 2.17 & 0.89 $\pm$ 0.01 \\
w/o $\mathcal{L}_{\sigma}$ & 33.13 $\pm$ 1.67 & 0.90 $\pm$ 0.02 \\
w/o $\mathcal{L}_{\text{interp}}^{\ast}$, $\mathcal{L}_{\sigma}$ & 32.76 $\pm$ 2.09 & 0.87 $\pm$ 0.02 \\
\ourinter{} & \textbf{33.52 $\pm$ 1.87} & \textbf{0.91 $\pm$ 0.01} \\
\bottomrule
\end{tabular}
\end{minipage}%
\hfill
\begin{minipage}{0.49\textwidth} 
\centering
\caption{MRI interpolation with different training sampling rates.}
\label{tab:interp_combined}
\fontsize{8pt}{9pt}\selectfont
\begin{tabular}{llcc}
\toprule
Frame & Method & PSNR $\uparrow$ & SSIM $\uparrow$ \\
\midrule
every & Linear & $32.70 \pm 1.90$ & $\mathbf{0.91 \pm 0.02}$ \\
2nd & Flow & $29.07 \pm 2.27$ & $0.86 \pm 0.04$ \\
 & \ourinter{} & $\mathbf{33.52 \pm 1.87}$ & $\mathbf{0.91 \pm 0.01}$ \\
\midrule
every & Linear & $30.44 \pm 1.93$ & $\mathbf{0.88 \pm 0.02}$ \\
3rd & Flow & $26.99 \pm 1.82$ & $0.82 \pm 0.03$ \\
 & \ourinter{} & $\mathbf{31.50 \pm 1.82}$ & $\mathbf{0.88 \pm 0.02}$ \\
\midrule
every & Linear & $28.10 \pm 2.14$ & $\mathbf{0.84 \pm 0.04}$ \\
5th & Flow & $25.08 \pm 1.56$ & $0.78 \pm 0.03$ \\
 & \ourinter{} & $\mathbf{28.93 \pm 1.90}$ & $0.83 \pm 0.03$ \\
\bottomrule
\end{tabular}
\end{minipage}

\end{table}









\noindent {\em Objective Function}
The total objective function combines losses from both tasks and temporal constraints:
$
\mathcal{L}_{\text{total}} = \mathcal{L}_{\text{img}} + \lambda_{\text{seg}}\mathcal{L}_{\text{seg}} + \lambda_{\sigma}\mathcal{L}_{\sigma}.$
The image loss $\mathcal{L}_{\text{img}}$ combines L1 reconstruction, Structural Similarity (SSIM), and $\mathcal{L}_{\text{interp}}^{\ast}$.
The segmentation loss $\mathcal{L}_{\text{seg}}$ minimizes the difference between rendered masks and ground truth masks. Finally, we regularize the time-spread parameter $\sigma_t$ of the Folded-Gaussians:
$
\mathcal{L}_{\sigma} = \frac{1}{N} \sum_{i=1}^N \left[ \max\left(\frac{2}{T} - \sigma_{t,i}, 0\right) + \max\left(\sigma_{t,i} - 1, 0\right) \right],
$
where $T$ is the number of training frames. This prevents Gaussians from being too temporally sparse or overly diffuse.

\noindent \textbf{Model Variants}
To evaluate the efficacy of our unified approach and address data heterogeneity, we introduce two specialized sub-models. In many clinical scenarios, datasets may be incomplete, providing only one modality—such as ultrasound sequences without corresponding segmentation masks. Consequently, \our{} is designed to function with a single head when specific annotations are unavailable:

\textbf{ I. \ourinter{} (Interpolation Only)}: This variant is optimized solely for the view synthesis and denoising of medical slices. It retains the shared geometry and the image-specific color module $\mathbf{c}^{\text{img}}$, while discarding the mask branch $\mathbf{c}^{\text{seg}}$. The objective reduces to $\mathcal{L}_{\text{img}} + \mathcal{L}_{\sigma}$, making it a suitable baseline for tasks where expert annotations are unavailable.

\textbf{ II. \ourseg{} (Mask Reconstruction Only)}: This variant focuses exclusively on anatomical surface extraction using only $\mathbf{c}^{\text{seg}}$. While it effectively reconstructs meshes from existing binary masks, it lacks the texture-based guidance provided by the original medical images. This single-head optimization is used to demonstrate that the absence of joint image-mask learning can lead to suboptimal results in low-contrast regions, where geometric continuity is more difficult to resolve.

\input{sections/experiments}




\section{Conclusions}
We introduced \our{}, the first Gaussian Splatting framework for sequential 3D medical imaging. By utilizing Folded-Gaussians and a shared-geometry multi-task architecture, our method enables robust frame interpolation and high-fidelity mesh reconstruction from sparse ultrasound and MRI data. Unlike implicit representations, \our{} offers an explicit, interpretable model that is efficient to train even with noisy inputs.
Extensive experiments show that \our{} outperforms classical and neural baselines in both quantitative accuracy and qualitative fidelity. These results position \our{} as a powerful tool for longitudinal disease assessment and precise anatomical visualization in clinical research.
\textbf{Limitations}  
As \our{} provides structural estimations rather than direct tissue measurements, its outputs involve inherent uncertainty and should serve as decision-support tools alongside expert clinical assessment.

%
%
\bibliographystyle{splncs04}
%





\end{document}

%% file: sections/experiments.tex
\section{Experiments}
\label{sec:experiments}
\noindent \textbf{Implementation Details}
\our{} is implemented in PyTorch and trained on an NVIDIA A100. Each run starts with 100k Folded-Gaussian components and uses a degree-7 polynomial for 
$f$ in Eq.~\eqref{eq:cond1} for multi-task training. $\lambda_{\text{seg}}$ and $\lambda_{\sigma}$ are set to 0.001 and 0.001 respectively. Training takes $\sim$ 20 minutes per object; rendering and mesh extraction take seconds.

\noindent \textbf{Interpolation Results}  
Interpolation is evaluated on the Skull-Stripped T1-Weighted MRI and TotalSegmentator \cite{d2024totalsegmentator} datasets. To simulate sparse sampling, a leave-frame-out protocol is used: training on every $n$-th slice ($n \in \{2,3,5\}$) and testing on the remaining slices. For baselines, we use linear approximation, which estimates intermediate slices by averaging neighboring slices, and optical flow, which computes voxel-wise motion between slices for guidance. As shown in Table~\ref{tab:interp_combined}, \our{} consistently outperforms linear interpolation and optical flow across PSNR and SSIM metrics. Qualitative results in Figure~\ref{fig:interp} confirm that \our{} maintains sharp anatomical boundaries, whereas baselines exhibit significant blurring or distortion.


\noindent \textbf{Mesh Reconstruction} For mesh reconstruction baselines, we use ISO-surface extraction, which converts segmentation masks into meshes via the marching cubes algorithm, Poisson reconstruction, which generates smooth surfaces by solving a 3D Poisson equation, and FUNSR, a self-supervised neural implicit method that learns signed distance functions from volumetric data to produce continuous 3D surfaces. We evaluate the mesh reconstruction task on three datasets. 
Table~\ref{tab:prostate_mesh_combined} summarizes results on the Prostate Ultrasound dataset. ISO-surface extraction produces jagged, noisy meshes, while Poisson reconstruction oversmooths surfaces and loses fine details. FUNSR achieves competitive fidelity, however, \our{} attains the lowest Chamfer and 95\% Hausdorff distances, indicating better geometric accuracy and anatomical preservation. The Dual-Task model (\our{}) achieves better performance than \ourseg{}, confirming the benefits of its joint optimization strategy.

In Figure~\ref{fig:prostate_mesh}, we highlight the high-quality meshes produced by \our{} on the Prostate Ultrasound and TotalSegmentator MRI dataset (heart, lung, and kidney). The gains are especially visible in a sparse sampling regime. 


Figure~\ref{fig:kidney_cancer} shows CTA-based 3D reconstructions for renal tumor characterization and aortic aneurysm assessment. For kidney cancer, the model provides precise 3D localization and orientation of the tumor and its relationship to residual parenchyma. This directly supports surgical planning, including selection between partial versus radical nephrectomy and choice of surgical approach (open vs. minimally invasive, e.g., robotic), and assessment of kidney-sparing feasibility, particularly in patients with limited renal reserve. For aortic aneurysms, the same CTA-based pipeline yields detailed 3D characterization of aneurysm location and morphology. This facilitates accurate planning of open versus endovascular repair, device selection and sizing, and prediction of technical challenges. In addition, the 3D reconstructions enable consistent longitudinal comparison across time points, improving assessment of aneurysm growth and timing of intervention, and ultimately supporting more individualized treatment decisions.


\noindent \textbf{Ablation Study}
We conduct an ablation study on the interpolation task using MRI brain volumes with a leave-frame-out strategy, omitting every second frame during training. We evaluate the effect of removing $\mathcal{L}_{\text{interp}}^{\ast}$ and $\mathcal{L}_{\sigma}$. As shown in Table~\ref{tab:ablation}, removing $\mathcal{L}_{\text{interp}}^{\ast}$ or $\mathcal{L}_{\sigma}$ decreases PSNR and SSIM. Excluding both components further degrades the results.